%% file: main.tex
\PassOptionsToPackage{table}{xcolor}
\documentclass[10pt,twocolumn,letterpaper]{article}

\usepackage[pagenumbers]{cvpr} % To force page numbers, e.g. for an arXiv version
\input{preamble}

\definecolor{cvprblue}{rgb}{0.21,0.49,0.74}
\usepackage[pagebackref,breaklinks,colorlinks,citecolor=cvprblue]{hyperref}
\usepackage[table]{xcolor}  % For coloring table cells
\usepackage{graphicx}
\usepackage{booktabs}
\usepackage{epsfig}
\usepackage{amsmath}
\usepackage{amssymb}
\usepackage{lipsum}
\usepackage{bm}
\usepackage{multirow}
\usepackage{colortbl}
\usepackage{arydshln}

\makeatletter
\@namedef{ver@everyshi.sty}{}
\makeatother

\usepackage{amsfonts} %% <- also included by amssymb
\DeclareMathSymbol{\shortminus}{\mathbin}{AMSa}{"39}

\newcommand{\PSNR}{PSNR $\uparrow$}

\newcommand{\SSIM}{ SSIM$\uparrow$}
\newcommand{\MRAE}{ SAM$\downarrow$}
\newcommand{\RMSE}{ RMSE$\downarrow$}
\newcommand{\FPS}{ FPS$\uparrow$}

\def\redc{\bf\cellcolor[HTML]{FF999A}}
\def\orangec{\it \cellcolor[HTML]{FFCC99}}
\def\yellowc{\cellcolor[HTML]{FFF8AD}}

\def\paperID{PISSOFF} % *** Enter the Paper ID here
\def\confName{CVPR}
\def\confYear{2024}

\title{HyperGS: Hyperspectral 3D Gaussian Splatting}

\author{
Christopher Thirgood\\
CVSSP, University of Surrey\\
{\tt\small c.thirgood@surrey.ac.uk}
\and
Oscar Mendez\\
CVSSP, University of Surrey\\
{\tt\small o.mendez@surrey.ac.uk}
\and
Erin Ling\\
CVSSP, University of Surrey\\
{\tt\small chao.ling@surrey.ac.uk}
\and
Jon Storey\\
I3D Robotics, Kent, UK\\
{\tt\small jstorey@i3drobotics.com}
\and
Simon Hadfield\\
CVSSP, University of Surrey\\
{\tt\small s.hadfield@surrey.ac.uk}
}

\begin{document}
\twocolumn[{%
\renewcommand\twocolumn[1][]{#1}%
\maketitle
    \captionsetup{type=figure}
    %\vspace{-9mm}
    \includegraphics[width=\textwidth]{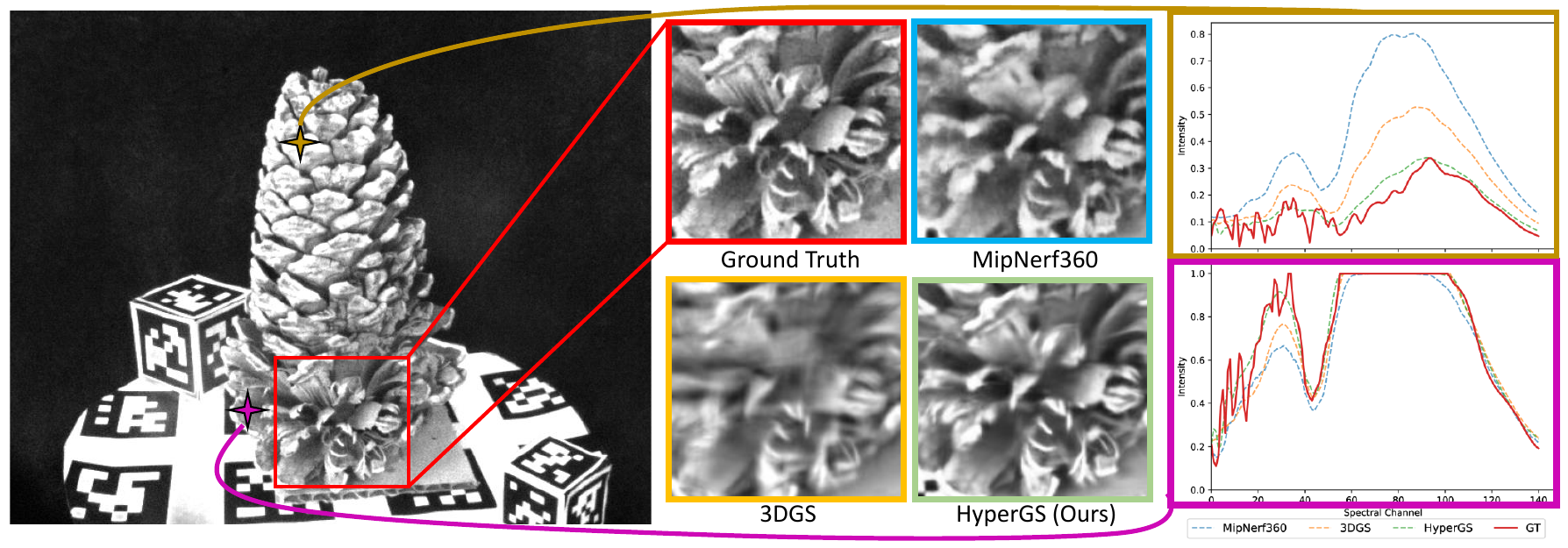} %\vspace{-2mm}
    \centering
    \vspace{-5mm}
    \hfill\caption{This image represents a novel hyperspectral image at the 70th channel of a 141-channel image predicted from the top three models, HyperGS (Ours), 3DGS and MipNerf360 for the hyperspectral novel view synthesis task. Two random pixel reconstructions taken from the novel view with corresponding border color for its plot and pixel location in the image represented by the stars.}
    \label{fig:teaser}
    \hfill \vspace{0mm}
    %\captionof{figure}{Test caption}
}]

\input{sec/0_abstract}  
% \vspace{-5mm}
\input{sec/1_intro}
\input{sec/2_litrev}
\input{sec/3_preliminary}

\input{sec/4_method}
\input{sec/5_experiments}
\input{sec/6_conclusion}
{
    \small
    \bibliographystyle{ieeenat_fullname}
    \bibliography{main}
}

\input{sec/X_suppl}

\end{document}

%% file: preamble.tex
\usepackage[dvipsnames]{xcolor}

%% file: sec/0_abstract.tex
\begin{abstract}
\vspace{-1mm}
We introduce HyperGS, a novel framework for Hyperspectral Novel View Synthesis (HNVS), based on a new latent 3D Gaussian Splatting (3DGS) technique. 
Our approach enables simultaneous spatial and spectral renderings by encoding material properties from multi-view 3D hyperspectral datasets. 
HyperGS reconstructs high-fidelity views from arbitrary perspectives with improved accuracy and speed, outperforming currently existing methods. 
To address the challenges of high-dimensional data, we perform view synthesis in a learned latent space, incorporating a pixel-wise adaptive density function and a pruning technique for increased training stability and efficiency. 
Additionally, we introduce the first HNVS benchmark, implementing a number of new baselines based on recent SOTA RGB-NVS techniques, alongside the small number of prior works on HNVS. 
We demonstrate HyperGS's robustness through extensive evaluation of real and simulated hyperspectral scenes with a \textbf{14db} accuracy improvement upon previously published models.
% This work marks a significant step forward in efficient multi-view 3D geometry synthesis for hyperspectral modeling.
\end{abstract}

%% file: sec/1_intro.tex
\section{Introduction}
\label{sec:intro}
Recent advancements in Novel View Synethesis (NVS), particularly with implicit Neural Radiance Field modeling (NeRFs) \cite{mildenhall2020nerf} and explicit Gaussian models (3DGS)\cite{3dgs}, have dramatically improved the fidelity of synthetically rendered views for conventional RGB images. 
However, RGB imaging is fundamentally limited, as it cannot capture the detailed material properties or non-visible scene characteristics critical for deeper scene understanding.

Hyperspectral imaging addresses these limitations by capturing a continuous spectrum of light across narrow bands for each pixel, providing valuable material properties and subtle spectral signatures. 
This modality is crucial for applications such as remote sensing, medical diagnostics, environmental monitoring, and robotics \cite{raspectloc}, where spectral information is paramount. 
The demand for Hyperspectral Novel View Synthesis (HNVS) is driven by the need for accurate, real-time spectral and spatial modeling.
However, synthesizing novel views from hyperspectral data presents significant challenges due to its high dimensionality and the requirement for spectral consistency across different perspectives for each pixel.

Previous works, such as HS-NeRF \cite{Chen2024_hyperspectral_nerf}, have explored HNVS by adapting NeRF architectures to hyperspectral data. 
However, these methods suffer from unstable training dynamics, slow rendering times, and inefficiencies in handling high-dimensional.
% NeRF-based models, with their large parameter counts, struggle to efficiently process the many spectral channels in hyperspectral data. 
% The first multi-view hyperspectral dataset provided by HS-NeRF is a valuable contribution, yet the method's limitations highlight the need for more robust solutions.

3DGS has recently emerged as a promising approach for fast NVS in the RGB domain, offering efficient, high-fidelity rendering of 3D scenes through point-based representations. 
By projecting 3D Gaussians onto the image plane, 3DGS enables smooth, continuous representations of scene geometry while preserving fine surface details. 
However, despite its efficiency and rendering quality, 3DGS has not yet been successfully extended to accommodate high-dimensional data types. 
In our benchmark, traditional 3DGS fails to achieve consistent results against NeRF baselines for HNVS (Section \ref{sec:experiments} and Section \ref{sec:scannet} of the supplementary materials). 
% This highlights the challenge of fast simultaneous spatial and spectral reconstruction for explicit modeling approaches like 3DGS. 
HyperGS aims to address these challenges. 
% by using a learned latent space and enhancements to the original 3DGS algorithm, integrating hyperspectral material information to enable more stable, high-quality spectral and spatial renderings.
In summary, the contributions of this paper are:
\begin{enumerate}
    \item We present the first method that successfully integrates view-dependent hyperspectral material information with 3DGS for high-quality HNVS.
    \item We introduce an adaptive density control and global pruning process that leverages hyperspectral signatures for efficiency alongside a hyperspectral SFM process to stabilize the modeling.
    \item A comprehensive benchmark of hyperspectral NeRF approaches and adaptations of classical RGB-NVS models for HNVS.
\end{enumerate}

%% file: sec/2_litrev.tex
\section{Literature review}
\label{sec:litreview}

\subsection{Hyperspectral 3D reconstruction}
Efforts to extend 3D reconstruction to multispectral and hyperspectral domains include works like Zia et al. \cite{Zia2015_hyperspectral_3d_reconstruction} and Liang J et al. \cite{Liang_3D_plants_hyper}, which employ point clouds and key point descriptors for structure-from-motion. 
These methods enhance 3D reconstruction by improving point matching but often result in sparse, noisy point clouds with inadequate occupancy information for multi-view stereo. Shadows and lighting variations further degrade geometric quality. A faster approach by Zia et al. \cite{Zia_hyper_on_mesh} instead skips rasterization by instead mapping hyperspectral images onto preprocessed 3D meshes.

\subsection{NeRFs}
NeRFs have gained significant attention \cite{Dellaert20blog_nerf_explosion} since Mildenhall et al. \cite{mildenhall2022cvpr_raw-nerf} introduced the technique.
% Deep learning is used to generate high-quality 3D reconstructions by mapping viewpoints to color radiance and volume density.
NeRFs rely on classical volume rendering techniques for scene synthesis.
Recent advances have sought to adapt NeRFs for spectral and hyperspectral imaging. SpectralNeRF \cite{Li2023_spectral_nerf} incorporates a spectrum attention mechanism (SAUNet) for realistic low-channel multi-spectral rendering under variable lighting. However, the reliance on attention mechanisms introduces computational overhead, making long-range spectra difficult to capture. Spec-NeRF \cite{Li2023_spec_nerf} takes a more efficient approach by adapting TensoRF \cite{tensorf} for reconstruction, with a seperate MLP estimating the camera’s spectral sensitivity, focusing on computationally efficient spectral scene reconstructions.
HS-NeRF \cite{Chen2024_hyperspectral_nerf} targets hyperspectral novel view synthesis by learning sinusoidal position encoding for each channel, allowing it to interpolate between viewpoints. This approach struggles with full-spectrum fidelity due to its lack of end-to-end training. 
HS-NeRF also released two multi-view hyperspectral datasets, using two different hyperspectral cameras with varying signal-to-noise ratios.
These, NeRF-based methods remain limited by their large parameter counts and instability when handling high-dimensional hyperspectral data alongside slower inference times.

\subsection{3DGS}
% Recent advancements in 3DGS have enhanced point cloud efficiency and quality which in the hyperspectral domain is key to improved accuracy in rendered images. 
% However, 3DGS’s reliance on user-defined thresholds often leads to uneven textures and memory inefficiencies, particularly when handling high-dimensional data like HSI.
Advancements in 3DGS improve point cloud efficiency for hyperspectral imaging but rely on discrete thresholds, causing uneven textures and memory issues with high-dimensional HSI data.
VDGS \cite{Malarz2023_gaussian_splatting_VDGS} employs a hybrid NeRF neural model for color and opacity, improving spectral reconstruction for HSI. However, this model still heavily depends on the 3DGS prediction and only uses opacity estimation from the MLP. Scaffold-GS \cite{Lu2023_scaffold_gs} introduces compressed representations to smooth surfaces but, like 3DGS, it struggles with artifacts in sparse regions. Mip-Splatting \cite{Yu2024MipSplatting} addresses rendering quality by applying 3D smoothing and a 2D Mip filter, yet it remains constrained by view-frustum-based Gaussian selection, limiting its effectiveness in dynamic scenes.
Challenges in Gaussian initialization, especially due to poor SfM data, are tackled by RAIN-GS \cite{jung2024relaxing_rain_gs}, which incorporates adaptive optimization. InstantSplat \cite{Fan2024_instantsplat} improves stability by using simpler point provisioning alongside the use of dust3r \cite{wang2023dust3r} which offers better SFM than COLMAP but has difficulty in processing large amounts of images. Meanwhile, Bulo \cite{Bulo2024_revising_densification} proposes pixel-level error-based densification to ensure consistent quality, and GaussianPro \cite{cheng2024gaussianpro} extends pixel-wise loss methods to better align 3D Gaussian normals during rendering, albeit at the cost of longer training times.
% HyperGS integrates a latent spectral exploration to improve error handling and an enhanced pixel-wise densification process for improved spatial modelling.
EfficientGS\cite{Liu2024_efficientgs} and LightGaussian\cite{Fan2023_lightgaussian} reduce model size and computational load through pruning methods but risk losing important visual details, especially when over-decimation occurs without proper ranking thresholds\cite{Liu2024_efficientgs}. These methods often prioritize storage efficiency over rendering performance, overlooking challenges in real-time optimization.
Although no previous works have explored the adaptation of Gaussian Splatting to Hyperspectral data, our experiments show that a naive adaptation struggles to model fine details.
HyperGS solves these issues by operating in a lower-dimensional learned latent space.

%% file: sec/3_preliminary.tex
\section{Preliminaries}
\label{preliminaries}

3DGS, developed by Kerbl et al. \cite{3dgs}, represents 3D environments using \(N\) Gaussian primitives. Each Gaussian is parameterized by its position \(x_i \in \mathbb{R}^3\) in world coordinates, a scaling vector \( \mathbf{S}_i \in \mathbb{R}^3\) defining the size of the Gaussian along each axis, a rotation quaternion \( \mathcal{R}_i \in \mathbb{R}^4\) defining its orientation, opacity \(\sigma_i \in \mathbb{R}\), which is controlled by a sigmoid function, and appearance features \(f_i \in \mathbb{R}^k\) to represent view-dependent RGB signals.
The core of 3DGS lies in transforming these Gaussians from 3D world space into the 2D image plane for rendering. To do this, we first define the 3D covariance matrix of each Gaussian in world space as:
\begin{equation}
{
\Sigma_i = R_i S_i^2 R_i^T,
}
\end{equation}
where \(R_i\) is the rotation matrix derived from the quaternion \(\mathcal{R}_i\), and \(S_i\) is the scaling matrix derived from the vector \(\mathbf{S}_i\).

To splat the Gaussian onto the 2D image plane, we transform the covariance matrix using the viewing transformation matrix \(W\) and the Jacobian matrix \(J\), which handles the projection onto the camera’s image plane as
\begin{equation}
{
\hat{\Sigma}_i = J W \Sigma_i W^T J^T.
}
\end{equation}
accounting for the perspective distortion and view transformation.
When rendering to a particular pixel, the opacity \(\alpha_i\) of Gaussian $i$ is computed as:
\begin{equation}
\alpha_i = 1 -  \exp(- \delta_i^T \hat{\Sigma}_i^{\shortminus 1} \delta_i),
\end{equation}
where \(\delta_i\) is the distance between the pixel and the projected center of the Gaussian in 2D. 

The transmittance \(T_i\), which represents the cumulative transparency along the ray up to the \(i\)-th Gaussian. It is computed as:
\begin{equation}
{
T_i = \prod_{j=1}^{i-1} (1 - \sigma_i\alpha_j) .
}
\end{equation}
The final color \(C\) of each pixel, $p$ is obtained by blending the colors of all Gaussian's projected onto that pixel:
\begin{equation}
{
C(p) = \sum_{i \in N} T_i \alpha_i c_i ,
}
\end{equation}
where \(p\) is the coordinate position in the image, \(c_i\) is the color of the \(i\)-th Gaussian, derived from its color appearance coefficients \(f_i\).
In summary, this process transforms and projects the Gaussian primitives from 3D space into 2D image space and computes the final pixel colors through opacity and transmittance blending.

%% file: sec/4_method.tex
\section{Methodology}
\label{sec:Method}
\begin{figure*}
    \centering
    \includegraphics[width=\linewidth]{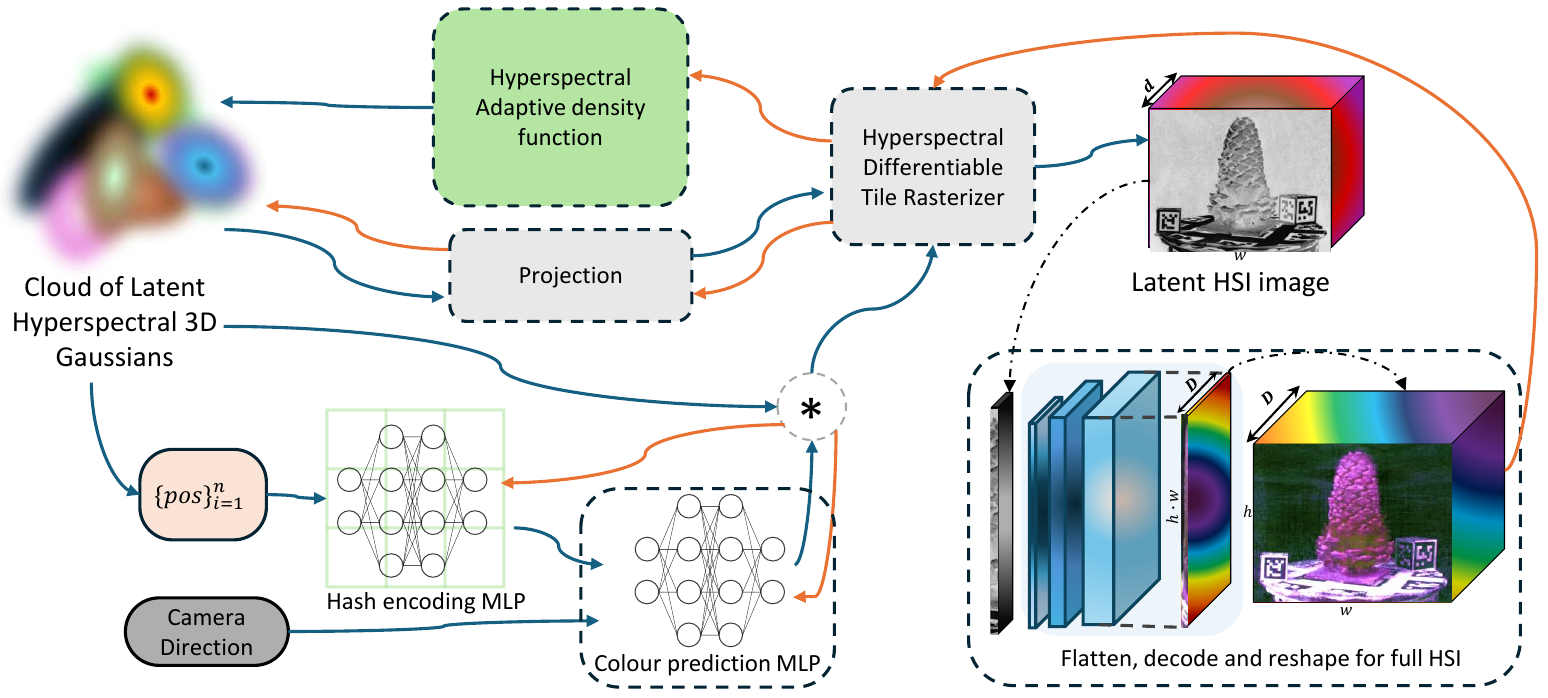}
    \caption{Visual system diagram of our approach. Blue lines indicate the operational flow, while orange lines represent the gradient flow. Our novel modified latent hyperspectral adaptive density function operates within a latent space provided by a frozen autoencoder, which is also responsible for decoding the final images. Latent space novel views are generated through a combination of the latent hyperspectral Gaussian point cloud and a NeRF-style MLP. Gradients are updated based on the decoded images.}
    % We first reproject a greyscale SFM point cloud onto LHSI from a pre-trained autoencoder. The 3DGS system explores an autoencoder's latent space with a spatially-aware MLP providing anisotropic view-dependent information and then decodes the predicted images back to full hyperspectral fidelity. The full reconstructed images are used for higher detailed loss computation during training. We also provide a densification protocol and apply Gaussian pruning using a pixel-wise top-k spectral similarity scoring method.}
    \label{fig:sysdiagram}
    \vspace{-2mm}
\end{figure*}
HyperGS aims to provide a lightweight, fast rasterization solution to HNVS that is robust to different hyperspectral cameras' sensitivity functions with outstanding accuracy. Our system diagram is seen in Figure \ref{fig:sysdiagram}.

\begin{figure}[t]
	\centering
	\includegraphics[width=\columnwidth]{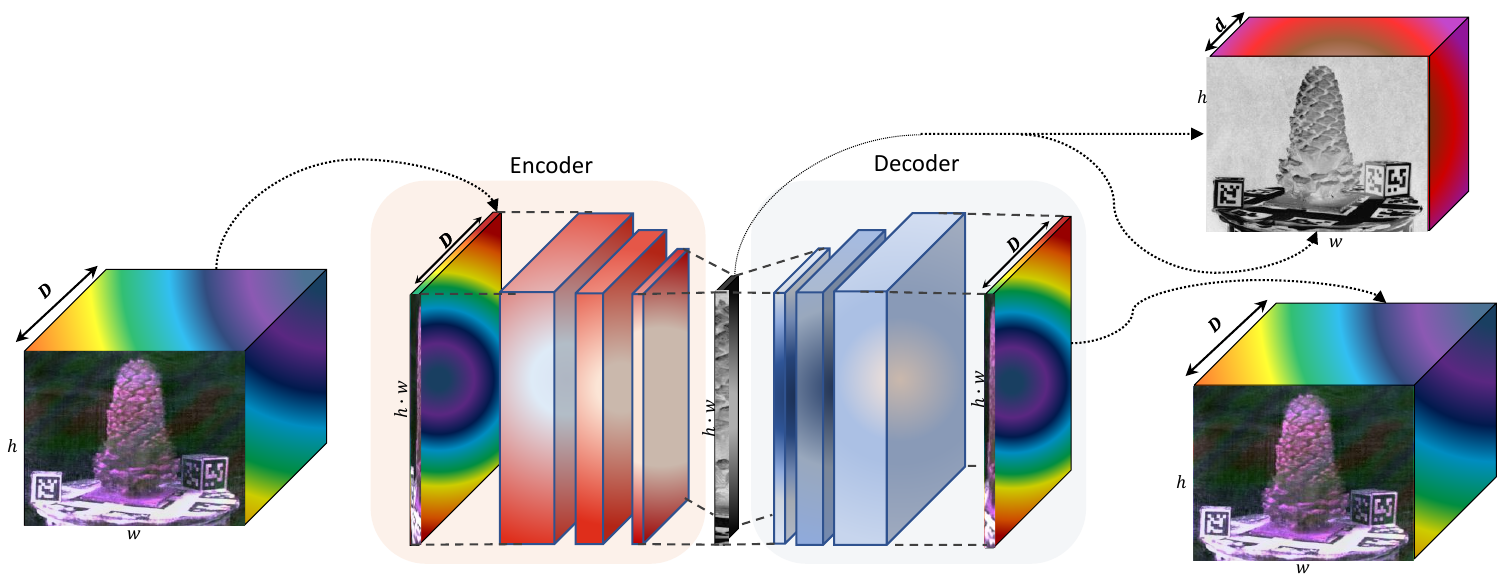}
	\caption{Our channel-wise convolutional AE model learns LHSI space representation of the scene. The decoder is only used in training and inference for the 3DGS system after the preprocessing of the dataset is finished.}
	\label{fi:Teacher}
 \vspace{-2mm}
\end{figure}
\subsection{Hyperspectral Compression} 
\label{sec:compression}
To address the challenges of high-dimensional optimization, we adopt a novel latent space exploration method for the hyperspectral data. 
The latent space of a pre-trained autoencoder (AE) serves as an exploratory space for our 3DGS system. This approach reduces the computational load during 3DGS optimization by providing a compact, lower-dimensional target. Additionally, the latent space bounds the error by encapsulating the spectral sensitivity of the hyperspectral camera for each channel, enhancing the accuracy and reliability of novel view reconstruction in the spectral domain. 
During testing, the latent space 3DGS viewpoint estimations are decoded using the frozen AE to produce novel HSIs from which gradient flow will be calculated.

The compression network is a fast convolutional AE. Both sides are symmetrical and is built using a series of 1D convolutional layers and Squeeze-Excitation (SE) blocks that operate across the spectral dimension of the images. The SE blocks benefit the performance by weighting features between layers for improved spectral discrimination. The model is trained on the pixel level of the scenes dataset. The encoder compresses the high-dimensional hyperspectral data into a latent representation via max-pooling layers, while the decoder reconstructs the original spectral information from this compressed form via upsampling. The architecture omits skip connections to ensure the decoder can function independently during training and testing to decode the LHSI data produced by the 3DGS system. For a detailed layout of the network architecture, please refer to Fig. \ref{fi:Teacher}.

The AE minimizes the loss \( L_{ae} \), defined as:
\begin{equation}
    L_{ae} = L_{Huber}(C^*(p), Dec(Enc(C^*(p)))) ,
\end{equation}
where $C^*$ is the ground truth pixel spectrum.
The Huber loss provides a smooth optimization process throughout training, as it handles outliers effectively while maintaining smooth gradients. 
Since each hyperspectral dataset includes varying levels of signal-to-noise ratios, this approach helps improve robustness against noise-related issues.

\subsection{Latent Hyperspectral 3D Gaussian Splatting}
\label{sec:adaptions}
Each Gaussian in the splatting process is assigned a latent spectral signature \( \mathbf{f}_i \in \mathbb{R}^m \), where \( m \) represents the latent space dimensionality. 
Inspired by many previous RGB techniques \cite{Lu2023_scaffold_gs, lee2024deblurring, yang2024specGaussian, malarz2023Gaussian} an MLP conditioned on view-directional hash encoding is used to predict anisotropic opacity and spectral color information for each Gaussian. 
This allows view-dependent spectral effects to be modelled upon specific bands. 
This signature is mapped directly to the pixel values using the decoder model defined in Section~\ref{sec:compression}. 
% Specifically, the latent vector \( \mathbf{s}_i \) can be decoded to produce the hyperspectral prediction \( \hat{\mathbf{h}}_i \) for each Gaussian.
% The decoder reconstructs the full hyperspectral image by decoding the latent spectral signatures assigned to each Gaussian.
% In the splatting process, the latent hyperspectral signatures \( \mathbf{s}_i \) are projected from the image plane into the 3D space, carrying the latent spectral information through the rendering pipeline. 
% Opacity adjustments depend on the viewing direction, which in turn affects the visibility of different Gaussians, resulting in dynamic color changes in the rendered image. This enhances the model's ability to represent hyperspectral data under varying lighting conditions. 
For camera $d$, the MLP $\mathcal{F}$ predicts view-dependent spectral effects \( \Tilde{f}_{i,d} \) and opacity \( \Tilde{\sigma}_{i,d} \) effects for each Gaussian. Specifically, the MLP takes the centre \( \mathbf{m}_i \) of each Gaussian and the view direction as inputs:
\begin{equation}
[\Tilde{f}_{i,d}, \Tilde{\sigma}_{i,d}]  = 
\mathcal{F}_{v}(h(\mathbf{m}_i, \mathbf{d}); \Theta),
\end{equation} 
where \(h\) is the hash encoding of the inputs while \( \Tilde{\Theta} \) denotes the MLP parameters similar to that of I-NGP \cite{mueller2022instant} and MipNerf360 \cite{barron22cvpr_mipnerf360}. 

These view-dependent spectral and opacity effects are multiplied with those stored in the Gaussian Cloud, leading to the updated volumetric rendering equations with transmittance being defined as:
\begin{equation}
{
T_{i,d} = \prod_{j=1}^{i-1} (1 - \alpha_j \sigma_i \Tilde{\sigma}_{i,d}).
}
\end{equation}
The final latent signature \(\hat{C}\) of each pixel is obtained by blending the colors of all Gaussians projected onto that pixel $p$:
\begin{equation}
{
\hat{C}(p,d) = \sum_{i \in N} T_{i,d} \alpha_i f_i \Tilde{f}_{i,d} .
}
\end{equation}
It is worth emphasizing that these view-dependent spectral effects are applied within the learned latent space before decoding. This helps maintain a low computational complexity while minimizing outliers.

A decoding operation is then performed to provide the full fidelity prediction from the system using the decoder from Section \ref{sec:compression}.
\begin{equation}
    C(p,d) = Dec( \hat{C} (p,d)) .
\end{equation}

The final rendering ensures that the spectral integrity is maintained throughout the entire process, which is critical for accurately reconstructing material properties. 
By leveraging the latent space, HyperGS provides a more meaningful and efficient spectral representation of the scene, than the original full hyperspectral image.
The original loss proposed in 3DGS for spatial and geometric consistency in rendered images uses a weighted DSSIM and L1 loss. 
However, this type of loss have been shown to lead to underfitting for HSI in other fields. This is because an L1 loss can produce extreme errors for sensitive spectral bands, destabilizing training.
To address this, we employ a weighted loss combining Charbonnier loss and cosine similarity to account for both spectral quality and spatial consistency. The cosine similarity provides a measured angular distance between vectors, making it ideal for comparing spectra.
To provide an overall consistency in the spatial and spectral domain we keep the SSIM score in the training from the original 3DGS system. 
The per pixel HyperGS training loss \( L_d(p) \) is:
\begin{equation}
    L_d(p) = (1-\lambda)
    (\beta L_{CB}(p)
    + L_{CS}(p) )
    + \lambda L_{SSIM}(p),
\end{equation}
where \(\beta\) weighs the spectral loss balancing the cosine similarity and charbonnier loss.
Thus the total training loss across all pixels and views is:
\begin{equation}
    L(C, C^*) = \sum_{p=1}^{P} \sum_{d \in V}  L_d(p),
\end{equation}
where \( C \) is the 3DGS systems network prediction after decoding the entire latent image, \( C^*\) is the ground truth hyperspectral image, and $V$ is the set of training view directions.

\subsubsection{Initialisation}
\label{sec:initialisation}
\begin{figure}
    \centering
    \includegraphics[width=\linewidth]{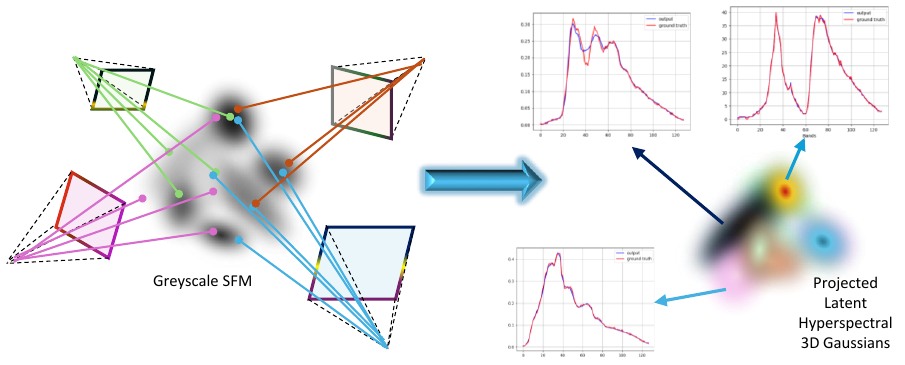}
    \caption{Visualization of our re-projection protocol for initializing the SFM point cloud. We estimate the SFM from grayscale channel slices of the hyperspectral image scene with COLMAP. Then, using the average of all views, we re-project each point into LHSI from provided by our AE, providing an optimal initialization of latent spectral color for the 3DGS system.}
    \label{fig:repro}
\end{figure}
Since there is no dedicated hyperspectral Structure-from-Motion (SfM) algorithm compatible with COLMAP, we first convert the hyperspectral images into grayscale images, \( I_{\text{gray}} \), by selecting the spectral channel with the highest foreground intensity variance to preserve key features for the SfM process. Our process is visualized in Figure \ref{fig:repro}.
Using these grayscale images and their corresponding camera projection matrices for each view \( d \), we apply COLMAP to generate a sparse 3D point cloud. Each 3D point \( \mathbf{X} = (X, Y, Z, 1)^T \) in world coordinates is related to its pixel coordinate \( \mathbf{p}_d = (x_d, y_d, 1)^T \) in the image plane via the camera projection \( \mathbf{K}_d\mathbf{E}_d \):
\begin{equation}
    \mathbf{p}_d = \mathbf{K}_d\mathbf{E}_d \mathbf{X},
\end{equation}
where \(\mathbf{K}_d\) are the camera intrinsics and \(\mathbf{E}_d\) the extrinsics.
After generating 3D points and recovering camera poses, we re-project the points into the LHSIs. Each 3D point \( \mathbf{X} \) is linked to pixel coordinates \( \mathbf{p}_d \) across all views. We then initialize the Gaussian cloud with one Gaussian centered on each 3D point \( \mathbf{X} \), with spectral signature \( \mathbf{f}_i \) computed by averaging the latent hyperspectral signatures \( H_d(y_d, x_d) \) across all views \( d \):
\begin{equation}    
f_i = \frac{1}{|V|} \sum_{v=1}^{V} \hat{C}^*_d(\mathbf{p}_d),
\end{equation}
where \( \hat{C}^*_d(\mathbf{p}_d) \) is the latent hyperspectral signature at pixel p for view \( d \), and \( V \) is the set of training views. This averaging ensures the spectral information is captured robustly.
%By combining re-projection with spectral averaging, we initalise the Gaussian cloud with appropriate latent hyperspectral information.
% This integration into the SfM pipeline enables 3D point clouds with detailed spectral signatures for HyperGS. However, sparse point clouds can lead to suboptimal model outcomes.

\subsubsection{Latent Hyperspectral Densification}
% \begin{figure}
%     \centering
%     \includegraphics[width=0.8\linewidth]{figures/images/densification_diagram_cropped.pdf}
%     \caption{Illustration of the densification process in HyperGS. The depth-scaling function reduces the influence of near-camera Gaussians by scaling their contribution based on depth relative to the scene radius. The splitting/cloning decision is based on the gradient magnitude across viewpoints, ensuring accurate hyperspectral scene reconstruction with minimal artifacts.}
%     \label{fig:pgradient}
%     \vspace{-2mm}
% \end{figure}
As the number of color channels in HSI data increases, visual discontinuities become more prevalent. Additionally, our initial SFM reconstruction is less dense compared to standard RGB 3DGS. To address these issues, we incorporate an advanced densification method that enhances stability in both results and training. 
A key component of 3DGS involves determining whether a Gaussian should be split or cloned based on the gradient magnitude of the Normalized Device Coordinates (NDC) across various viewpoints.
%To this end the influence radius \(R_{v}^{g}\) for each Gaussian \(g\) under viewpoint \(d\) is computed using the Gaussian's 2D projected covariance matrix \(\Sigma'_i\). The equation for the influence radius is:
%\begin{equation}
%R_{d}^{g} = 3 \times \left( \frac{\lambda_1^{g,d} + \lambda_2^{g,d}}{2} + \sqrt{\left(\frac{\lambda_1^{g,d} + \lambda_2^{g,d}}{2}\right)^2 - \lambda_1^{g,d} \lambda_2^{g,d}} \right),
%\end{equation}
%where \(\lambda_1^{g,d}\) and \(\lambda_2^{g,d}\) are the eigenvalues of the 2D projected covariance matrix \(\Sigma'_i\). These eigenvalues describe the extent of the Gaussian in the 2D image plane after applying the viewing transformation and projection matrices. 
%This influence radius \(R_{v}^{g}\) encapsulates 99\% of the Gaussian's probability distribution and is used to filter the Gaussian's influence.
%The 3DGS system typically decides whether to split or clone Gaussians based on the gradient magnitude of the NDC across different viewpoints. 
However, in sparse regions, this can cause artefacts as large Gaussians are highly visible in many viewpoints, leading to inconsistent splitting. This challenge is exacerbated in the hyperspectral domain of 3DGS, where the larger number of channels and highly variable value ranges make it harder to set effective thresholds. To mitigate this, we scale the gradient by the square of the depth relative to the scene’s radius, which reduces the influence of Gaussians near the camera.
More formally, we define the splitting score as:
% \begin{equation}
%     \mathcal{S}(g_i) = 
%     \sum_{v=1}^V \sum_{p=1}^P
%     h(v,i)
%     \sqrt{
%     \left(\frac{\partial L_v(p)}{\partial x_i}\right)^2 +
%     \left(\frac{\partial L_v(p)}{\partial y_i}\right)^2
%     },
% \end{equation}
\begin{equation}
\!    \mathcal{S}(g_i) = 
\!    \sum_{d \in V} \sum_{p=1}^P
\!    h(d,i)
\!    \smash{\sqrt{
\!    \left(\frac{\partial L_d(p)}{\partial x_i}\right)^2 
\!    +
\!    \left(\frac{\partial L_d(p)}{\partial y_i}\right)^2
\!    }},
\end{equation}
where the depth-scaling function is:
\begin{equation}
    h(d,i) = \left( \frac{|\mathbf{E}_d \mathbf{X}_i|}{\beta_\text{field} \times R} \right)^2,
\end{equation}
\( \beta_\text{field} \) is a tunable parameter, and \( R \) is the scene's radius, based on the largest distance between any pair of cameras.

The Gaussian $g_i$ is split or cloned if $\mathcal{S}(g_i) > \theta_q$.
%\begin{equation}
%\frac{\mathcal{S}_q}{\sum_{j=1}^{V_q} p_j^q} > \theta_q.
%\end{equation}
This approach accounts for depth-based scaling and pixel contributions, leading to better hyperspectral scene reconstruction with fewer artifacts near the camera. However, this more expressive densification can result in an excessive number of Gaussians in scenes with high-frequency details (e.g., the ``pinecone'' scene in Figure \ref{fig:teaser}, to stabilize this we deploy a global pruning strategy.

\subsection{Global Hyperspectral Gaussian Pruning} 
\label{sec:pruning}
\begin{figure}
    \centering
    \includegraphics[width=\linewidth]{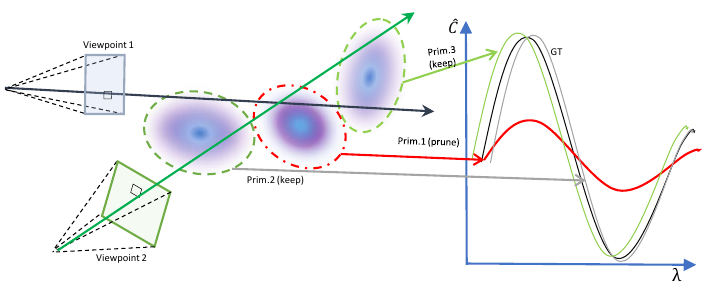}
    \caption{
    % Visualization of our pixel-wise pruning process. Gaussians with similarity scores below a threshold are pruned, such as the red-circled Gaussian with poor similarity to the black ground truth spectrum.
    Visualization of our pixel-wise pruning. Gaussians with poor similarity scores below a threshold are pruned, such as the red-circled Gaussian and the black ground truth spectrum.
    }
    \label{fig:pruning}
    \vspace{-2mm}
\end{figure}
In HyperGS, we deploy pixel-wise pruning \cite{Niedermayr2023_compressed_Gaussian_splatting} to preserve the spectral integrity of latent hyperspectral images (LHSIs) while lowering the overall number of primitives to improve the scene quality, model size, and spectral gradient descent. 
Unlike cross-view pruning methods, which evaluate Gaussians across multiple viewpoints and leads to over-pruning \cite{Girish2023_eagles, Liu2024_efficientgs}, our approach assesses each Gaussian’s contribution at the pixel level. This approach retains fine spectral details specific to each pixel. 
% Cross-view pruning, while effective for RGB data, tends to overlook pixel-specific spectral information, making it less effective for hyperspectral imaging.
For each Gaussian \( g_i \), pruning is based on its pixel-wise importance $\mathcal{I}$ across all viewpoints and pixels. We compute the importance for every combination of Gaussian, pixel and viewpoint as the spectral difference between the ground truth hyperspectral value \( \hat{C}^*_d(p) \) and the decoded Gaussians LHSI, with the pruning score defined as:
% \begin{equation}
% \text{Acc}(g_i, p) = \left( 1 - \text{mean}\left( \lVert C(p) - \hat{C}(p) \rVert_1 \right) \right) \cdot \alpha_i \cdot T_i,
% \end{equation}
\begin{equation}
\mathcal{I}[g_i, p, d] = \left( 1 - \left| C^*_d(p) - Dec(f_i) \right| \right) \alpha_i T_i,
\end{equation}
where \( T_i \) is the accumulated transmittance of Gaussian \( g_i \). Including \( \alpha_i \) and \( T_i \) ensures that only Gaussians contributing significantly in visibility and spectral accuracy are retained, and also ensures that the score is 0 for Gaussians that do not overlap the given pixel.

Following recent works, we do not prune Gaussians based on the average, or the total pixel-wise importance score. Instead, we retain all Gaussians within the top-K importance ranking for any pixel.
More formally the filtered Gaussian cloud $\mathcal{G}$ is
\begin{equation}
\mathcal{G} = \left\{ g_i | \exists (p,d), 
%\text{ such that }
\text{Rank} ( g_i, \mathcal{I}[:,p,d] < \tau_p \right )\},
\end{equation}
where ``$\text{Rank}$'' is a function that returns the rank of a given Gaussian within the slice of the score tensor for that pixel and view.

%If the total pruning score for \( g_i \) summed across all viewpoints falls below a threshold \( \tau_p \), the Gaussian is pruned:
%\begin{equation}
%\text{Acc}(g_i, p) < \tau_p.
%\end{equation}
%The threshold \( \tau_p \) is set based on 10\% of the AE’s validation performance. We found this provides a balanced pruning efficiency and maintains spectral fidelity. Pruning is applied twice during densification to remove redundant Gaussians as more detailed representations emerge, maintaining computational efficiency without sacrificing spectral detail.

%% file: sec/5_experiments.tex
\section{Experiments}
\label{sec:experiments}
We first evaluate HyperGS using the HS-NeRF datasets \cite{Chen2024_hyperspectral_nerf}. These datasets differ in signal-to-noise ratios, channel lengths, and the number of images per scene, providing a comprehensive test of model performance. The Bayspec dataset \cite{Chen2024_hyperspectral_nerf} contains around 360 images per scene (3 scenes total), while the SOP dataset \cite{Chen2024_hyperspectral_nerf} has around 40 images per scene (4 scenes total).
Due to the lack of continuous HSI datasets for non-object-centric scenes, we also evaluate HyperGS on a synthetic dataset curated from ScanNetv200 \cite{dai2017scannet} (Section \ref{sec:scannet} in the supplementary materials, where we replace each semantic label with distinct hyperspectral signatures as seen in our supplementary materials).
The baselines we have implemented for the HNVS problem include conventional 3DGS (with our reprojected SFM initialisation \ref{sec:initialisation}), traditional NeRF models adapted for hyperspectral data (NeRF \cite{mildenhall2020nerf}, MipNeRF \cite{barron2021mipnerf}  MipNeRF-360 \cite{barron22cvpr_mipnerf360}, Nerfacto \cite{nerfstudio} and TensorF \cite{tensorf} (Spec-NeRF), and the only existing HNVS method HS-NeRF \cite{Chen2024_hyperspectral_nerf}.

All experiments were conducted on an NVIDIA A100 80GB GPU, as MipNeRF360 requires significantly more VRAM than the other methods, which could run on a NVIDIA 3090 GPU. Using the A100 also facilitated accurate tracking of training times across all methods.
We assess the performance of HyperGS and competing methods using rigorous metrics that capture both accuracy and spectral fidelity: PSNR, SSIM, Spectral Angle Mapping (SAM), RMSE. 
We use a 90\% training and 10\% test split according to the HS-NeRF dataset.

\subsection{Real HSI}
\label{sec:RealHSIQuant}
\begin{figure*}[t]
    \centering
    \includegraphics[width=\linewidth]{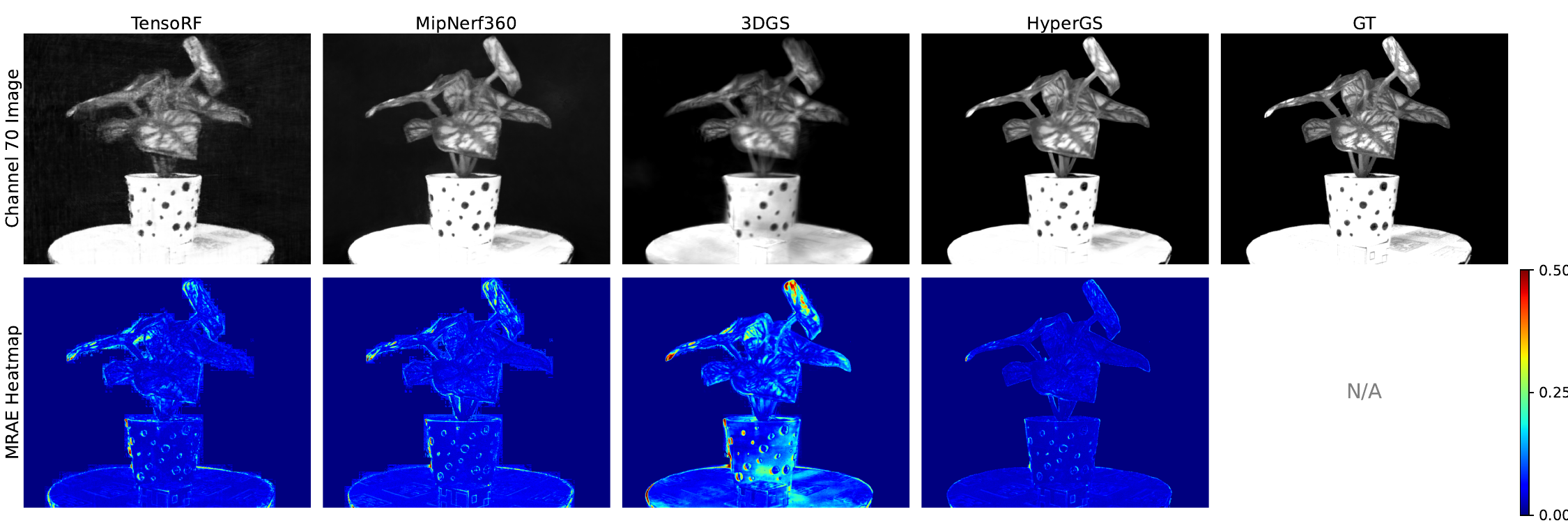}
    % \vspace{-2mm}
    \vspace{-8mm}
    \caption{Visualisation of the top 4 methods for frame 51 of 359 for the Caladium plant scene from the Bayspec dataset. The top row shows the 70th channel of the 141 channel image predicted, the bottom row provides a raw pixel-wise error heatmap of the scene.}
    \label{fig:Fakeplant2_1_heatmaps}
    \vspace{-4mm}
\end{figure*}
\begin{figure*}
    \centering
    \includegraphics[width=\linewidth]{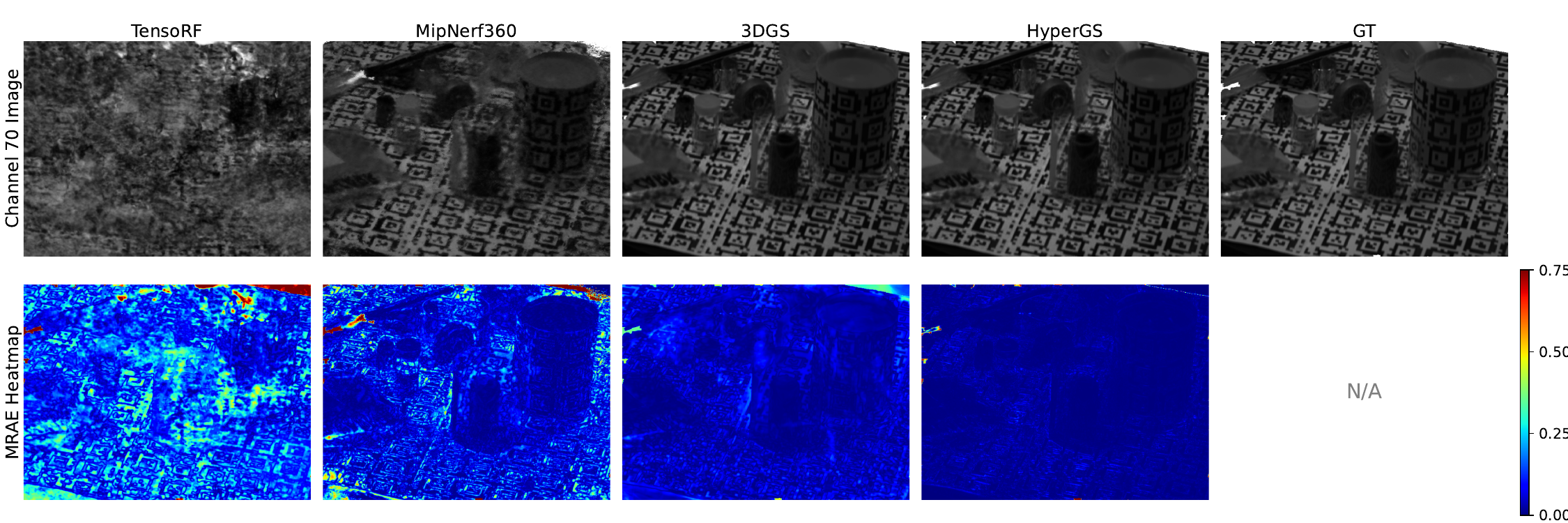}
    \vspace{-8mm}
    \caption{Visualisation of the top 4 methods for frame 31 of 49 for the Tools plant scene from the SOP dataset. The top row shows the 70th channel of the 128 channel image predicted, the bottom row provides a raw pixel-wise error heatmap of the scene.}
    \label{fig:sop_tools_heatmaps}
    \vspace{-4mm}
\end{figure*}
% \begin{figure*}[htp]
%     \centering
%     \begin{subfigure}{\textwidth}
%         \centering
%         % \caption{Three random pixel reconstructions from test frame 151 of the bayspec Anacampseros scene.}
%         \includegraphics[width=\textwidth]{figures/images/pixel_reconstruction_comparison_frame_00051_test5_fake2_1_no_key.pdf}
%         \label{fig:fake2_1_pixels}
%         \vspace{-5mm}
%     \end{subfigure}
%     \begin{subfigure}{\textwidth}
%         \centering
%         % \caption{Three random pixel reconstructions from test frame 31 of the SOP origami scene.}
%         \includegraphics[width=\textwidth]{figures/images/pixel_reconstruction_comparison_frame_00031_test4_tool_00041_3.pdf}
%         \label{fig:origami_pixels}
%     \end{subfigure}
%     \vspace{-8mm}
%     \caption{Pixel reconstruction comparisons for two different scenes: (a) bayspec Anacampseros scene (frame 151) and (b) SOP origami scene (frame 31).}
%     \label{fig:combined_pixels}
%     \vspace{-5mm}
% \end{figure*}
\begin{figure*}[htp]
    \centering
    \includegraphics[width=\linewidth]{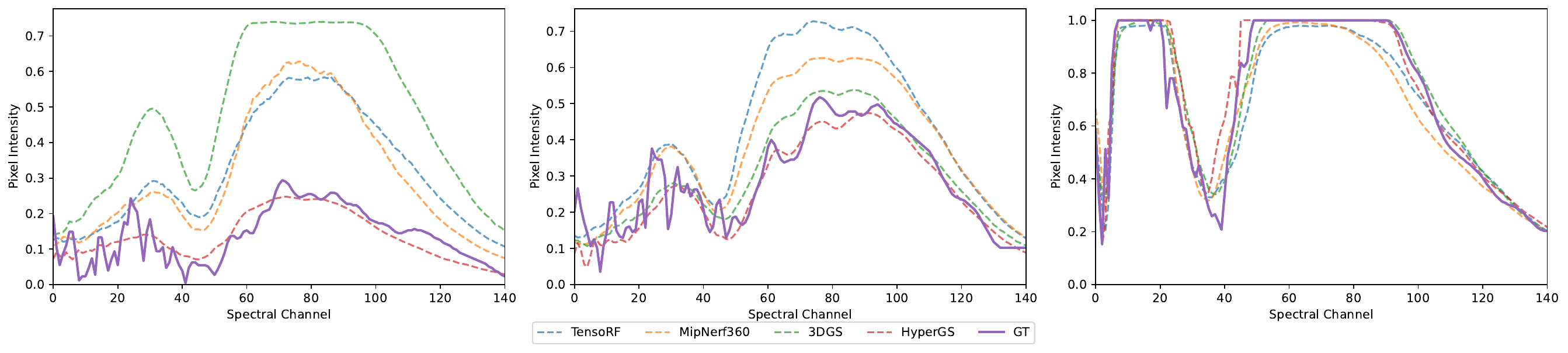}
    % \vspace{-5mm}
    \vspace{-8mm}
    \caption{Three random pixel reconstructions taken from test frame 151 of the bayspec Anacampseros scene.}
    \label{fig:fake2_1_pixels}
    \vspace{-4mm}
\end{figure*}
\begin{figure*}[htp]
    \centering
    \includegraphics[width=\linewidth]{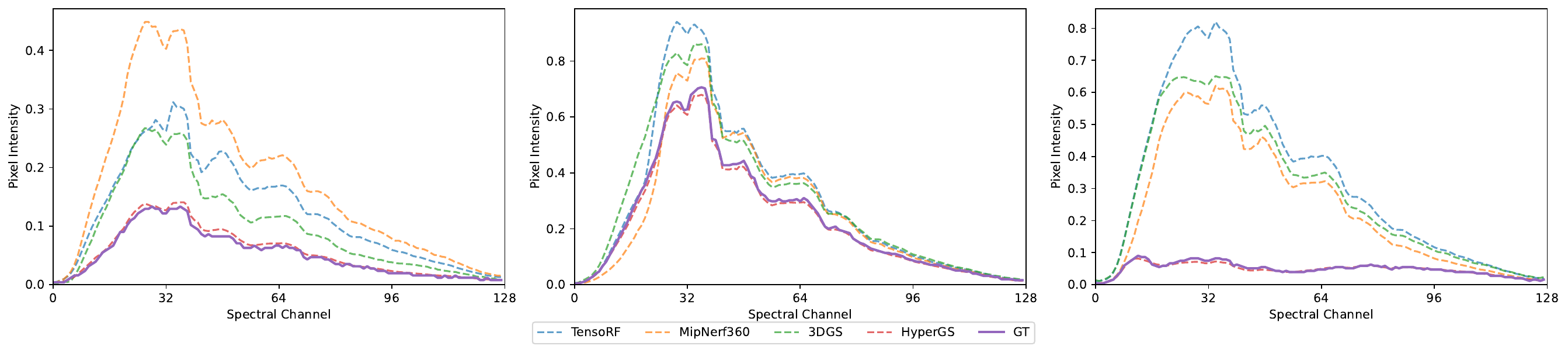}
    \vspace{-8mm}
    \caption{Three random pixel reconstructions from test frame 31 of the SOP origami scene.}
    \label{fig:origami_pixels}
    \vspace{-3mm}
\end{figure*}
We compare two HNVS datasets from HS-NeRF with varying spectral channels and noise. The SOC710-VP (SOP) camera provides high spatial (696 × 520) and spectral (128 channels) resolution but has poor temporal resolution due to long exposure times, covering 370-1100nm. The BaySpec GoldenEye camera offers similar spatial (640 × 512) and spectral (141 channels) resolution but captures faster, introducing more grain and noise.
Baseline performance results for these cameras are shown in Table \ref{tab:bayspec_avg} and Table \ref{tab:sop_avg}. HyperGS consistently outperforms existing methods in terms of quality on unseen images across all scenes. 

The BaySpec dataset particularly highlights the advantages of our architecture. The autoencoder’s role in generating bounded errors within the latent space enables HyperGS to effectively manage noise by averaging spectra and decoding within well-defined 1D feature structures. This approach leads to more accurate reconstructions than conventional end-to-end training methods used by the other models. This is further illustrated in Figure \ref{fig:Fakeplant2_1_heatmaps} where the size of the object and specular reflections in the spectral channels are predicted more accurately than with other baseline methods.

Similar to its performance on natural NeRF tasks, MipNeRF360 performed the best among NeRF methods for HNVS. We attribute its performance to its use of positional encoding and the efficient scene representation using tensor decomposition initially employed by TensorF providing more targeted spectral predictions.
Our densification and pruning techniques further refine HyperGS's performance by effectively filtering poor spectral representations. % as illustrated in Figure\ref{fig:visHSNeRF}. 
% Furthermore, the 3DGS model's FPS performance is ten times better or more than that of NeRF models in HNVS.
Moreover, the Surface Optics Pro (SOP) dataset yields far better results in the HNVS task for 3DGS approaches. Due to the high number of frames present in the BaySpec camera scenes, the NeRF approaches adapt well to this abundance of data and can provide far better and more stable representations of the scene given the volume of viewpoints. This volume makes predicting the noisier spectra that the BaySpec camera provides easier than 3DGS. In contrast, although the SOP datasets may have smoother spectra, the number of training viewpoints is far smaller.
We notice a drop-off in performance for NeRF approaches in this scenario, as they struggle to understand the scene scale and perform spectral reconstruction due to the lack of data. However, 3DGS approaches deliver far superior performance on this dataset by providing smoother interpolation of the data and avoiding complex optimization. This is outlined by Figure \ref{fig:sop_tools_heatmaps} highlighting the intense error heatmaps from NeRF approaches compared to 3DGS and HyperGS.

HyperGS, however, outperforms all methods across all scenes and camera datasets, suggesting that it provides a far more robust and accurate performance than both 3DGS and NeRF methods for HNVS.
% , while only sacrificing a small amount of FPS performance compared to 3DGS.
For visualizations of random pixels and spectral images, please refer to Figures \ref{fig:fake2_1_pixels} and then \ref{fig:origami_pixels} which showcases HyperGS's superior pixel reconstruction capabilities compared to other benchmarked methods.
% \begin{table}
%   \centering
%   \small
%   % \scriptsize
%   Surface Optics Datasets
%   \input{figures/tables/average_sop}\\[0.4em]
%   BaySpec Datasets\\
%   \input{figures/tables/average_bayspec}
%   \caption{Quantitative results using the Bayspec and surface optics pro datasets against separate hyperspectral methods and baseline NeRF and 3DGS.}
%   \vspace{-5mm}
%   \label{tab:bayspec_avg}
% \end{table}
\begin{table}
  \centering
  % \small
  % \scriptsize
  BaySpec Datasets
  \resizebox{\columnwidth}{!}{
  % Surface Optics Datasets
  % \input{figures/tables/average_sop}\\[0.4em]
  \input{figures/tables/average_bayspec_avg_no_fps}
  }
  \caption{Quantitative results using the Bayspec datasets against separate hyperspectral methods and baseline NeRF and 3DGS (best \textbf{bold}, second best \textit{italic}).}
  % \vspace{-3mm}
  \label{tab:bayspec_avg}
\end{table}
\begin{table}
  \centering
  % \small
  % \scriptsize
  Surface Optics Datasets
  \resizebox{\columnwidth}{!}{
  \input{figures/tables/average_sop_no_fps} %\\[0.4em]
  % BaySpec Datasets\\
  % \input{figures/tables/average_bayspec}
  }
  \vspace{-1mm}
  \caption{Quantitative results using the SOP datasets against separate hyperspectral methods and baseline NeRF and 3DGS.}
  \vspace{-5mm}
  \label{tab:sop_avg}
\end{table}

\subsection{Ablation Study} 
\label{sec:ablation}

To evaluate the effectiveness and impact of various components in our HyperGS model, we conducted a series of ablation experiments. These experiments help to understand the contribution of individual components and choices in our model's design.

% As shown in Table \ref{tab:able_methods} the results highlight the essential features of HyperGS, in providing better results with each added feature to our method improving the overall results. Table \ref{tab:able_methods}, highlights how the change in quality from the latent space can drastically change the overall performance of the whole model. From this, we suggest keeping the channel space high enough to keep reconstruction performance satisfactory but also keeping the size small enough for the 3DGS system to capture less rigid latent space details. 
% \subsubsection{Method analysis} 
As shown in Table \ref{tab:able_methods}, each new feature added to the baseline 3DGS model improves HNVS performance. The student-teacher architecture boosts spectral reconstruction the most while the proposed densification provides greater overall details captured. With the latent space learning and positional encoding, the performance of HyperGS gets a significant boost in performance too.
\begin{table}
  \centering
  % \small
  % \scriptsize
    \resizebox{\columnwidth}{!}{
  % Simulated Scannet Dataset
  \input{figures/tables/ablation_methods}%\\[0.5em]
  }
  \vspace{-1mm}
  \caption{Adding each new feature to 3DGS significantly improves the model's average performance for the HNVS task in the bayspec dataset while creating a more stable number of primitives (N.Prim) in the point cloud.}
  \label{tab:able_methods}
  \vspace{-5mm}
\end{table}
% \vspace{-10mm}
Please refer to our supplementary materials for additional ablations, including analyses of HyperGS performance, pruning strategy scoring functions, autoencoder performance with varying data and latent sizes, the global pruning strategy, and system speed bottlenecks.\looseness=-1
% \subsubsection{Pruning strategy}
% Choosing the appropriate pruning score function is crucial for maintaining spectral fidelity. We tested several functions—including L1, L2, Huber, SAM, and mean average error (MAE) as shown in Table \ref{tab:abla_pruning_f}. We also provide the results without the pruning to highlight the positive effect it has on the system. The L1-Norm loss performed best, balancing detail preservation and model simplicity by penalising large deviations while keeping the structure intact. SAM preserved angular relationships but resulted in worse channel intensity preservation. L2 led to over-pruning, degrading reconstruction quality in regions with complex spectral features due to smoothing of the latent spectra when global pruning was activated.
% \begin{table}
%   \centering
%   \small
%   \scriptsize
%   % Simulated Scannet Dataset
%   \input{figures/tables/ablation_pruning_f} % Removed the \\ here
%   % \vspace{0.5em} % Use \vspace for spacing instead of \\[0.5em]
%   \caption{Ablation performance difference using difference pruning functions for latent hyperspectral Gaussians in the bayspec dataset.}
%   \label{tab:abla_pruning_f}
%   \vspace{-2mm}
% \end{table}

%% file: figures/tables/average_bayspec_avg_no_fps.tex
% \scriptsize
\begin{tabular}{l cccc } 
  \toprule
  \multicolumn{1}{c}{\multirow{2}{*}{Method}} & \multicolumn{4}{c}{Average Results}  \\  % Adjust the multirow alignment
          & \PSNR & \SSIM & \MRAE & \RMSE \\
  \cmidrule(lr){1-1} % Rule under the Method column
  \cmidrule(lr){2-5} %\cmidrule(lr){5-5}
  
  NeRF         & 23.35 & 0.6061 & 0.0440 & 0.0687  \\
  MipNeRF      & 22.75 & 0.5947 & 0.0435 & 0.0776  \\
  TensoRF   & \yellowc24.66 & \yellowc0.6482 & \yellowc0.0501 & \yellowc0.0587 \\
  Nerfacto     & 19.12 & 0.5866 & 0.0551 & 0.1174  \\
  MipNerf360     & \orangec26.53 & \orangec0.7442 & \orangec0.0280 & \orangec0.0476 \\
  HS-NeRF      & 19.82 & 0.6714 & 0.0534 & 0.1071 \\
  \midrule
  3DGS         & 22.91 & 0.6321 & 0.1335 & 0.0810 \\
  HyperGS    & \redc27.11 & \redc0.7804 & \redc0.0254 & \redc0.0440\\
  \bottomrule
\end{tabular}

%% file: figures/tables/average_sop_no_fps.tex
% \scriptsize
\begin{tabular}{l cccc } 
  \toprule
  \multicolumn{1}{c}{\multirow{2}{*}{Method}} & \multicolumn{4}{c}{Average Results} \\  % Adjust the multirow alignment
          & \PSNR & \SSIM & \MRAE & \RMSE \\
  \cmidrule(lr){1-1} % Rule under the Method column
  \cmidrule(lr){2-5} %\cmidrule(lr){6-6}
  
  NeRF         & 10.89 & 0.5905 & 0.0625 & 0.3479  \\
  MipNeRF      & 12.53 & 0.5481 & 0.0568 & 0.3198  \\
  TensorF     & 13.00 & 0.5696 & 0.0595 & 0.2780  \\
  Nerfacto     & \yellowc16.37 & \yellowc0.6986 & \yellowc0.0352 & 0.1601  \\
  MipNeRF360   &  12.28 & 0.6824 & 0.1369 & 0.2658  \\
  HS-NeRF      & 14.44 & 0.6165 & 0.2037 & \yellowc0.1953  \\
  \midrule
  3DGS         & \orangec28.58 & \orangec0.9627 & \orangec0.0301 & \orangec0.0478  \\
  HyperGS    & \redc30.51 & \redc0.9756 & \redc0.00415 & \redc0.0354  \\
  \bottomrule
\end{tabular}

%% file: figures/tables/ablation_methods.tex
\begin{tabular}{l ccccc} % 6 columns: 1 for ablation step, 5 for metrics including num primitives
  \toprule
  \multirow{2}{*}{Ablation Step} & \multicolumn{5}{c}{Average Results for the Bayspec dataset} \\ % Span over the metrics
   % Rule under the metrics
            & \PSNR & \SSIM & \MRAE & \RMSE & N.Prim $\downarrow$ \\
  \cmidrule(lr){1-1} \cmidrule(lr){2-6} % Rule under the Ablation Step column
  Base. 3DGS        & 22.91 & 0.6320 & 0.1335 & 0.0810  & 440k \\
  + Spec. SFM      & 23.05 & 0.6331  & 0.1310  & 0.0799 & 421k \\
  + Latent space AE    & 24.87 & 0.7101  & 0.0548  & 0.0602 & 500k \\
  + Densification      & \yellowc25.25 & \yellowc0.7356  & \yellowc0.0365  & \yellowc0.0548 & 1.3M \\
  + Pruning            & 25.17 & 0.7199  & 0.0374  & 0.0555 & \yellowc412k \\
  + View depenent MLP  & \orangec27.05 & \orangec0.7792  & \redc0.0253  & \orangec0.0443 & \redc309k \\
  + Custom L.Func      & \redc27.11 & \redc0.7801 & \orangec0.0254 & \redc0.0440 & \orangec310k \\
  \bottomrule
\end{tabular}

%% file: sec/6_conclusion.tex
\vspace{-1mm}
\section{Conclusions} \label{sec:conclusion}
In this paper, we introduced HyperGS, the first effective approach for hyperspectral novel view synthesis (HNVS). By encoding latent spectral data within Gaussian primitives and performing Gaussian splatting in a learned latent space, HyperGS achieves high-quality, material-aware rendering. Our adaptive density control and pruning techniques enable efficient handling of latent hyperspectral signatures in 3D Gaussian point clouds, resulting in stable training and superior accuracy compared to benchmarked methods.
For future work, we plan to extend HyperGS to provide a greater latent space decoding method that will encapsulate and correct not only spectral but spatial information for improved contextual detail.

%% file: sec/X_suppl.tex
\clearpage
\setcounter{page}{1}
\maketitlesupplementary
In this supplementary materials document, we present the complete results for all scenes in the Bayspec and SOP datasets in Section \ref{sec:Additional_Results}. Additionally in Section \ref{sec:scannet} we evaluate our performance on third dataset: a simulated ScanNetv200 dataset \cite{dai2017scannet}, where spectral replacements are applied to the target images. Furthermore, we provide ablation studies on various aspects of our approach, including the pruning strategy scoring function (Section \ref{sec:Pruning_strat}), the performance of the system's autoencoder with varying latent space dimensionality and the global pruning call frequency to the system (Section \ref{sec:pruning}).
% , and an investigation into the primary speed bottlenecks in our system (Section \ref{sec:performance_bottleneck}). 

We chose to ablate on the Bayspec dataset scenes in development, and we found larger improvements in optimising from the Bayspec dataset that translated into the others. This can be attributed to Bayspec's high-frequency information in most scenes, more noisy spectra, and general significance of images in the scenes.

% \begin{table*}[htp]
%   \centering
%   % \small
%   % \scriptsize
%   Surface Optics Datasets
%   \input{figures/tables/sop_main}\\[0.4em]
%   BaySpec Datasets\\
%   \input{figures/tables/bayspec_main}
%   \caption{Quantitative results using the HS-NeRF dataset against separate hyperspectral methods and baseline NeRF and 3DGS.}
%   % \vspace{-5mm}
%   \label{tab:full_hsnerf}
% \end{table*}
\begin{table*}[htp]
  \centering
  % \small
  % \scriptsize
  Surface Optics Datasets
  \resizebox{\textwidth}{!}{
  \input{figures/tables/sop_main}%\\[0.4em]
  }
  % BaySpec Datasets\\
  % \input{figures/tables/bayspec_main}
  \caption{Quantitative results using the HS-NeRF dataset against separate hyperspectral methods and baseline NeRF and 3DGS.}
  % \vspace{-5mm}
  \label{tab:full_SOP}
\end{table*}
\begin{table*}[htp]
  \centering
  % \small
  % \scriptsize
  % Surface Optics Datasets
  BaySpec Datasets\\
  \resizebox{\textwidth}{!}{
  \input{figures/tables/bayspec_main}
  }
  \caption{Quantitative results using the HS-NeRF dataset against separate hyperspectral methods and baseline NeRF and 3DGS.}
  % \vspace{-5mm}
  \label{tab:full_Bayspec}
\end{table*}

\begin{table}[htp]
  \centering
  % \small
  % \scriptsize
  Simulated Scannet Dataset
  \resizebox{\columnwidth}{!}{
  \input{figures/tables/average_scannet_no_fps}%\\[0.5em]
  }
  \caption{Quantitative results using the simulated hyperspectral scannet dataset against separate hyperspectral methods and baseline NeRF and 3DGS. Change colours to yellow, orange, red}
  \label{tab:scannetSim}
  % \vspace{-5mm}
\end{table}

\section{Per-scene quantitative results}
\label{sec:Additional_Results}
% \subsection{Bayspec and SOP datasets}
Tables \ref{tab:full_SOP} and \ref{tab:full_Bayspec} break down the results of the SOP and Bayspec datasets results of Tables \ref{tab:sop_avg} and \ref{tab:bayspec_avg} from the main paper, into metrics for each scene. Our method consistently enhances scene modeling performance on almost every combination of metric and scenarios. This is especially true in terms of improving the PSNR index for scene modeling. PSNR measures the extremism of outliers in the spectra of the predicted image which highlights the overall benfits of HyperGS over other methods. 
% We find that only in one of the scenes does Hyper
% Additionally, we stronglysuggest that readers watch the videos on the project page for a more direct understanding of how our approach surpasses 3DGS.

\section{Simulated Scannet dataset}
\label{sec:scannet}
Since no room-scale multi-view hyperspectral dataset currently exists, we propose simulating such a dataset using the ScanNetV2 dataset. To enhance the diversity and spectral richness, we utilize longer channel depths by incorporating downsampled Raman spectra from the open-source RRUFF dataset. Specifically, we select 200 suitable spectra (from their material description) from the RRuFF dataset and downsample them by a factor of 16, resulting in spectra with 228 channels. These spectra replace the segmentation labels in ScanNetV2 to generate a large number of simple simulated hyperspectral images. This approach provides valuable insights into the performance degradation of systems at high channel depths, even when a dataset includes a large number of viewpoints. Furthermore, to perform camera pose estimation from the colmap scenes, we downsampled the number of images and downscaled the resolutions of the image by 2 in order to get a working camera intrinsic estimation from the scene with COLMAP.

For fairness, we evaluate against the same baselines used throughout the mains paper, excluding MipNeRF360, as it was explicitly designed for ``turntable style'' 360-degree datasets. Table \ref{tab:scannetSim} summarizes the average results for four randomly selected scenes from the ScanNetV2 dataset.

The results consistently demonstrate that HyperGS is more robust and accurate than all other methods evaluated again. 
% Notably, HyperGS leveraged a latent space of only 29 channels and significantly outperformed 3DGS in this test.
While 3DGS also performed well compared to other baselines, its success can be attributed to the use of cleaner and less varied spectra, which was found to particularly benefit 3DGS in the SOP dataset. NeRF-based methods, as expected, were to perform better on ScanNet than on the SOP dataset due to the larger number of viewpoints available. However, 
%even though both methods were provided the benefitting factors of the real hyperspectral datasets,
HyperGS still provides greater results. We attribute this to other baselines performance degrading substantially when tasked with handling high channel counts, highlighting their limitations in hyperspectral scenarios. HyperGS overcomes these issues with the use of it's learned latent space.

We also provide the full set of quantitative results for the scannet dataset performance in Table \ref{tab:full_scannetSim}.
\begin{table*}
  \centering
  % \small
  % \scriptsize
  Simulated Scannet Dataset
  \input{figures/tables/scannet_main}%\\[0.5em]
  \caption{Quantitative results using the simulated hyperspectral scannet dataset against separate hyperspectral methods and baseline NeRF and 3DGS. Change colours to yellow, orange, red}
  \label{tab:full_scannetSim}
  % \vspace{-5mm}
\end{table*}
% We aim to show here a more qualitative comparison between our method and those referenced in \ref{sec:experiments}. Will provide additional results of several viewpoints and scenes.
% \begin{figure}
%     \centering
%         \includegraphics[width=\linewidth]{figures/images/hyperGS-teaser.png}
%         \includegraphics[width=\linewidth]{figures/images/hyperGS-teaser.png}
%         \includegraphics[width=\linewidth]{figures/images/hyperGS-teaser.png}
%         \includegraphics[width=\linewidth]{figures/images/hyperGS-teaser.png}
%         \includegraphics[width=\linewidth]{figures/images/hyperGS-teaser.png}
%     \caption{\textbf{PROTOTYPE IMAGE}.More images like this but for additional viewpoints and scenes and models}
%     \label{fig:enter-label}
% \end{figure}

\section{Additional ablation studies}
\label{sec:Ablation_extra}
% Here we provide results behind the reasoning's of several features and ablation parameters chosen in the HyperGS model.
In this section we aim to provide greater ablation studies on the bayspec dataset for HyperGS. We found that refining the technique on this dataset yielded greater results in the SOP dataset that we tested on as part of the main paper. 

\subsection{Pruning Strategy}
\label{sec:Pruning_strat}
% \subsubsection{Pruning strategy}
Choosing the appropriate pruning score function is crucial for maintaining spectral fidelity after the pruning is performed. We tested several functions—including L1, L2, Huber, SAM, and mean average error (MAE) as shown in Table \ref{tab:abla_pruning_f}. We also provide the results without the pruning to highlight the positive effect it has on the system (called `None' in the table). The L1-Norm loss performed best, balancing detail preservation and model simplicity by penalizing large deviations while keeping the structure intact. SAM preserved angular relationships but resulted in worse channel intensity preservation. L2 led to over-pruning, degrading reconstruction quality in regions with complex spectral features due to smoothing of the latent spectra when global pruning was activated. Hence in the final model we used an L1 loss in equation 17.
\begin{table}
  \centering
  % \small
  % \scriptsize
  \resizebox{\columnwidth}{!}{
  % Simulated Scannet Dataset
  \input{figures/tables/ablation_pruning_f} % Removed the \\ here
  % \vspace{0.5em} % Use \vspace for spacing instead of \\[0.5em]
  }
  \caption{Ablation performance difference using difference pruning functions for latent hyperspectral Gaussians in the bayspec dataset.}
  \label{tab:abla_pruning_f}
  % \vspace{-2mm}
\end{table}

\subsection{Autoencoder Training Strategy} 
We investigated the performance differences between an autoencoder trained individually for all the scenes in Bayspec versus a single autoencoder trained on all scenes within the dataset. 
We perform this experiment because all baselines are per-scene models and we aimed to provide a fairer test experiment. The single-scene AE approach is tailored to each scene's unique characteristics, potentially capturing finer details, while the general approach may benefit from broader exposure, improving robustness across different scenes. Our results, Table \ref{tab:ae_gen} show that the single AE provides better reconstruction quality in scenes with high variability, as it can specialize in scene-specific features. However, the general autoencoder, trained on all scenes, offers more consistent performance and reduced outliers across varied environments, albeit with a slight trade-off in specific scene detail and overall performance. In development, we found adding the MLP from Section \ref{sec:adaptions} takes a stronger role in providing better spectral reflections in the scene when added to the system, providing better latent spectral understanding.
\begin{table}[htp]
  \centering
  % \small
  % \scriptsize
  % Simulated Scannet Dataset
  \input{figures/tables/ablation_AE_dataset_V_scene}\\[0.5em]
  \caption{Ablation performance difference between using a general autoencoder trained on all scenes for each camera dataset against individual autoencoders trained per scene.}
  \label{tab:ae_gen}
\end{table}

% \subsection{Performance bottleneck} 
% \label{sec:performance_bottleneck}
% We highlight here some performance improvements for future work, namely we identify where the bottleneck lies in the rendering speed of HyperGS. Its clear that the pruning if effective as seen from \ref{tab:able_methods} however, due to the Teacher networks pixel wise compression, the speed of the complete hyperspectral rendering is still far slower than the 3DGS system in place.
% \begin{table}
%   \centering
%   \small
%   % \scriptsize
%   % Simulated Scannet Dataset
%   \input{figures/tables/ablation_speed}\\[0.5em]
%   \caption{\textbf{FAKE PLACEHOLDER RESULTS}. Identification where the bottleneck occurs in rendering. This table highlights that future work may look into faster rasterisation methods involving the autoencoder with computational techniques instead.}
%   \label{tab:abla_speeds}
% \end{table}

\subsection{Latent space ablation} 
\label{sec:latspace}
In this ablation, we tested the performance of HyperGS against the change in latent space size. Reducing latent space size can make the feature space more meaningful and compact. However, this necessitates additional AE layers, leading to a more rigid latent space, increased prediction errors and slower performance. 
%However, this leads to more layers being added to the AE and the latent space becoming more rigid and it shrinks leading to more errenous predictions.
In table \ref{tab:abla_latent}, we test against all of the datasets presented in the main and supplementary materials with ranging latent space sizes. We chose to do divisions of 4 and 6 of the full channel depth of the hyperspectral images to highlight the changes in performance when the latent space is reduced on the differing types of camera datasets. 
Interestingly, latent space performance varies significantly for the noisier Bayspec dataset, likely due to reduced expressiveness in handling noisy regions.
Whereas smoother hyperspectral data like that of SOP and scannet highlights the ability to comfortably transition to smaller channel sizes and in some cases outperform the division of 4 size used in the main paper.
To provide a fair and controlled experiment we use the division of 4 for all results in the main paper and the scannet results, since this provides the most consistent results.
\begin{table}
  % \centering
  % \small
  % \scriptsize
  % Simulated Scannet Dataset
  \resizebox{\columnwidth}{!}{

\input{figures/tables/ablation_latent_space}
  }
  \caption{Ablation performance for differing latent space sizes for the teacher model for the average metrics over all three datasets tested.}
  \label{tab:abla_latent}
  \vspace{-5mm}
\end{table}

\subsection{Pruning Frequency}
\label{sec:pruning_freq}
In this section, we aim to determine the optimal frequency of global pruning within the HyperGS approach. We evaluate three pruning strategies:
\textbf{Single Pruning During Densification:} Pruning once during the densification process allows the model to recover any Gaussians lost during pruning, if needed.
\textbf{Double Pruning During Densification:} Pruning twice during the densification process enables further reduction in the number of Gaussians but may impair accuracy if the model cannot achieve higher detail with fewer overall Gaussians.
\textbf{Hybrid Pruning:} Pruning once during densification and once after densification concludes, to assess the stability and safety of pixel-wise pruning effects.
As shown in Table \ref{tab:abla_pruning_f}, the best-performing method was to prune once during the densification procedure. This improvement may be attributed to the recovery of important information that can be reintroduced during densification. If global pruning is performed twice within the procedure, accuracy is further compromised, although the number of primitives decreases significantly. The downside of a single pruning call is that it leaves a larger number of primitives in the model.

Pruning after the densification process results in a substantial loss of accuracy, indicating that the information lost becomes unrecoverable and the overall output model is negatively affected.
The hybrid method provides results similar to post-densification pruning in terms of accuracy and the number of primitives. This demonstrates that the pruning method yields consistent pruning results even when called twice.
To achieve the best and most accurate results with the HyperGS system, we utilize a single global pruning call during the densification iterations for the Bayspec dataset.
\begin{table}
  \centering
  % \small
  % \scriptsize
  \resizebox{\columnwidth}{!}{
  % Simulated Scannet Dataset
  \input{figures/tables/ablation_pruning_freq} % Removed the \\ here
  % \vspace{0.5em} % Use \vspace for spacing instead of \\[0.5em]
  }
  \caption{Ablation study on performance differences using various pruning frequencies for the Bayspec dataset. 'In Densif.' refers to pruning within the densification procedure before 17.5K iterations. 'Post Densif.' refers to pruning at 17.5K iterations, and 'Hybrid' refers to pruning once before and once at 17.5K iterations.}
  \label{tab:abla_pruning_f}
  % \vspace{-2mm}
\end{table}

%% file: figures/tables/sop_main.tex
\scriptsize
\setlength{\tabcolsep}{3pt}
\begin{tabular}{l cccc cccc cccc cccc | c} 
  \toprule
  \multirow{2}{*}{Method} & \multicolumn{4}{c}{Rosemary} & \multicolumn{4}{c}{Basil} & \multicolumn{4}{c}{Tools} & \multicolumn{4}{c}{Origami} & \multirow{2}{*}[-1.25ex]{\FPS} \\  % Adjust the multirow alignment
           & \PSNR & \SSIM & \MRAE & \RMSE & \PSNR & \SSIM & \MRAE & \RMSE & \PSNR & \SSIM & \MRAE & \RMSE & \PSNR & \SSIM & \MRAE & \RMSE & \\
  \cmidrule(lr){1-1} % Rule under the Method column
  \cmidrule(lr){2-5} \cmidrule(lr){6-9} \cmidrule(lr){10-13} \cmidrule(lr){14-17} \cmidrule(lr){18-18}
  
  NeRF       
  %rosemary
  &   8.42
  &  0.7461
  &  0.0284
  &  0.3560 
  %basil
  &   9.91
  &  0.5534
  &  0.0769
  &  0.5256 
  %tools
  & 11.61 
  &  0.4962
  &  0.0610
  &  0.3018 
  %origami
  & 13.64 
  &  \yellowc0.5684
  &  0.0835
  &  0.2083 
  & 0.12 \\
  MipNeRF      
  %rosemary
  & 13.64*
  &  0.5684*
  &  1000*
  &  0.2083* 
  %basil
  & 10.11
  &  0.5878
  &  0.0728 
  &  0.5334
  %tools
  &  12.78 
  &  0.5213
  &  0.0598
  &  0.2781 
  %origami
  &  11.697
  &  0.5149
  &  0.0956
  &  0.2595 
  & 0.092 \\
  TensoRF   
  %rosemary
  &   12.1
  &  0.73351
  &  0.0212
  &  0.2662 
  %basil
  &   15.23
  &  0.5811
  &  0.0435
  &  0.3628  
  %tools
  &   11.697*
  &  0.5149*
  &  0.0956*
  &  0.2595*
  %origami
  &  12.98
  &  0.4488
  &  \yellowc0.0776
  &  0.2314 
  & 0.195 \\
  Nerfacto    
  %rosemary
  &  18.66
  &  0.8836
  &  0.0078
  &  0.1205
  %basil
  &  16.54
  &  0.7915
  &  0.0176
  &  0.1655 
  %tools
  &  16.254
  &  0.6135
  &  \yellowc0.0198
  &  0.1549  
  %origami
  &  \yellowc14.02
  &  0.5028
  &  0.0953
  &  \yellowc0.1993
  & \yellowc0.572 \\
  MipNerf360   
  %rosemary
  &  8.47
  &  0.7518
  &  0.0876
  &  0.3825
  %basil
  &  13.92
  &  \yellowc0.8584
  &  0.0497
  &  0.2035 
  %tools
  &  \yellowc16.80
  &  \yellowc0.7241
  &  0.0832
  &  0.1482 
  %origami
  &  9.93
  &  0.3951
  &  0.3271
  &  0.3288
  & 0.011 \\
  HS-NeRF     
  %rosemary
  &  \yellowc*18.60
  &  \yellowc*0.887
  &  \yellowc*0.0077
  &  \yellowc*0.1187 
  %basil
  &  \yellowc*16.81
  &  *0.771
  &  \yellowc*0.0172
  &  \yellowc*0.1587 
  %tools
  & *12.001 & *0.355 & *0.470 & *0.185 
  %origami
  &  10.359
  &  0.4530
  &  0.3197
  &  0.3188
  & 0.488 \\
  \midrule
  3DGS   
  %rosemary
  & \orangec25.56 & \orangec0.9695 & \orangec0.0028 & \orangec0.0534
  %basil 
  & \orangec21.19 & \orangec0.9385 & \orangec0.0101 & \orangec0.0897 
  %tools
  & \orangec29.13 & \orangec0.9596 & \orangec0.0165 & \orangec0.0391 
  %origami PSNR 38.46854782104492 SSIM 0.9833765029907227 RMSE 0.012825900688767433 MRAE 38.746971130371094 SAM 0.00034
  & \orangec38.46 & \orangec0.9833 & \orangec0.0003 & \orangec0.0128 
  & \redc79.0 \\
  HyperGS    
  %rosemary
  &  \redc26.77
  &  \redc0.9845
  &  \redc0.0021
  &  \redc0.0445
  %basil
  &  \redc25.30
  &  \redc0.9503
  &  \redc0.00514
  &  \redc0.0569
  %tools
  &  \redc30.86
  &  \redc0.9773
  &  \redc0.0091
  &  \redc0.0288 
  %origami
  &  \redc39.12
  &  \redc0.9906
  &  \redc0.0002
  &  \redc0.0114 
  & \orangec3.56 \\
  \bottomrule
\end{tabular}

%% file: figures/tables/bayspec_main.tex
\scriptsize
\begin{tabular}{l cccc cccc cccc | c} 
  \toprule
  \multirow{2}{*}{Method} & \multicolumn{4}{c}{Pinecone} & \multicolumn{4}{c}{Caladium} & \multicolumn{4}{c}{Anacampseros} & \multirow{2}{*}[-1.25ex]{\FPS} \\  % Adjust the multirow alignment
           & \PSNR & \SSIM & \MRAE & \RMSE & \PSNR & \SSIM & \MRAE & \RMSE & \PSNR & \SSIM & \MRAE & \RMSE & \\
  \cmidrule(lr){1-1} % Rule under the Method column
  \cmidrule(lr){2-5} \cmidrule(lr){6-9} \cmidrule(lr){10-13} \cmidrule(lr){14-14}
  
  NeRF 
  %pinecone
  &  22.82
  &  0.6113
  &  0.0446
  &  0.0728
  %fakeplant2_1
  &  23.12
  &  0.58348
  &  \yellowc0.0491
  &  0.0709
  %fakeplant1_2
  &  24.12
  &  0.6220
  &  0.0384
  &  0.0623
  & 0.13 \\
  MipNeRF      
  %pinecone
  & 21.45
  & 0.5738
  & \yellowc0.0410
  & 0.0856 
  %fakeplant2_1
  &  23.36
  &  0.5935
  &  0.0487
  &  0.0685
  %fakeplant1_2
  &  23.43
  &  0.6160
  &  0.0408
  &  0.0786
  & 0.090 \\
  TensoRF
  %pinecone
  & \yellowc24.12
  & \yellowc0.6454
  & 0.0593
  & \yellowc0.0625
  %fakeplant2_1
  &  \yellowc24.79
  &  0.6424
  &  0.0516
  &  \yellowc0.0577
   %fakeplant1_2
  &  \yellowc25.07
  &  0.6569
  &  0.0394
  &  \yellowc0.0558
  & 0.17 \\
  Nerfacto
  %pinecone
  & 15.36
  & 0.4935
  & 0.0707
  & 0.1709
  %fakeplant2_1
  &  20.67
  &  0.6208
  &  0.0529
  &  0.0945
  %fakeplant1_2
  &  21.32
  &  0.6423
  &  0.0417
  &  0.0867
  & \yellowc0.50 \\
  MipNeRF360     
  %pinecone
  & \orangec25.93
  & \orangec0.7355
  & \orangec0.0279
  & \orangec0.0507
  %fakeplant2_1
  &  \orangec26.93
  &  \orangec0.7371
  &  \orangec0.0332
  &  \orangec0.0461
    %fakeplant1_2
  &  \redc26.73
  &  \redc0.7601
  & \orangec0.0230
  &  \orangec0.0461
  & 0.010 \\
  HS-NeRF      
  & 20.07 
  & 0.581 
  & 0.0725 
  & 0.1521
  %fakeplant2_1
  &  19.084
  &  0.705
  &  0.0533
  &  0.0902
  %fakeplant1_2
  &  20.32
  &  \yellowc0.7260
  &  \yellowc0.0345
  &  0.0789
  & 0.47 \\
  \midrule
  3DGS 
  & 22.65 & 0.6039 & 0.0668 & 0.0819 
  & 23.50 & \yellowc0.7131 & 0.2889 & 0.0758 
  & 22.59 & 0.5786 & 0.0447 & 0.0853 
  & \redc78.1 \\
  HyperGS   
  & \redc27.0 
  & \redc0.7509 
  & \redc0.0309 
  & \redc0.0447 
  %fake2_1
  & \redc27.70 
  & \redc0.8354
  & \redc0.0271 
  & \redc0.0414
  %fake1_2
  & \orangec26.62 
  & \orangec0.7545
  & \redc0.0183 
  & \redc0.0460 
  & \orangec2.31 \\
  \bottomrule
\end{tabular}

%% file: figures/tables/average_scannet_no_fps.tex
% \scriptsize
\begin{tabular}{l cccc } 
  \toprule
  \multicolumn{1}{c}{\multirow{2}{*}{Method}} & \multicolumn{4}{c}{Average Results} \\  % Adjust the multirow alignment
          & \PSNR & \SSIM & \MRAE & \RMSE \\
  \cmidrule(lr){1-1} % Rule under the Method column
  \cmidrule(lr){2-5} %\cmidrule(lr){6-6}
  
  NeRF      & \yellowc15.85 & \yellowc0.7200 & \yellowc0.1509 & \yellowc0.1742 \\
  MipNeRF   & 14.45 & 0.7180 & 0.1700 & 0.2094 \\
  TensoRF   & 7.353 & 0.3522 & 0.8201 & 0.4599 \\
  Nerfacto  & 7.928 & 0.3896 & 0.6711 & 0.4520 \\
  HS-NeRF   & 7.363 & 0.3258 & 0.4107 & 0.4649 \\
  \midrule
  3DGS      & \orangec20.618 & \orangec0.8224 & \orangec0.06421 & \orangec0.1140 \\
  HyperGS   & \redc25.12 & \redc0.8805 & \redc0.04602 & \redc0.05833 \\

  \bottomrule
\end{tabular}

%% file: figures/tables/scannet_main.tex
\scriptsize
\setlength{\tabcolsep}{3pt}
\begin{tabular}{l cccc cccc cccc cccc | c} 
  \toprule
  \multirow{2}{*}{Method} & \multicolumn{4}{c}{0000-00} & \multicolumn{4}{c}{0009-00} & \multicolumn{4}{c}{0645-01} & \multicolumn{4}{c}{0703-01} & \multirow{2}{*}[-1.25ex]{\FPS} \\  % Adjust the multirow alignment
           & \PSNR & \SSIM & \MRAE & \RMSE & \PSNR & \SSIM & \MRAE & \RMSE & \PSNR & \SSIM & \MRAE & \RMSE & \PSNR & \SSIM & \MRAE & \RMSE & \\
  \cmidrule(lr){1-1} % Rule under the Method column
  \cmidrule(lr){2-5} \cmidrule(lr){6-9} \cmidrule(lr){10-13} \cmidrule(lr){14-17} \cmidrule(lr){18-18}
  
  NeRF       
  %0000
  &   \yellowc14.18
  &  0.6912
  &  \yellowc0.1113
  &  \yellowc0.2097
  %0009
  &  \yellowc16.53
  &  \yellowc0.7298
  &  \yellowc0.0978
  &  \yellowc0.1657 
  %0645
  & \yellowc16.28 
  &  \yellowc0.6691
  &  0\yellowc.1372
  &  \yellowc0.1650 
  %0703
  & \yellowc16.39
  &  0\yellowc.7889
  &  \yellowc0.2572
  &  \yellowc0.1564 
  & 0.070 \\
  MipNeRF      
  %0000
  & 13.47
  &  \yellowc0.7062
  &  0.1266
  &  0.2280 
  %0009
  & 14.47
  &  0.7076
  &  0.1212 
  &  0.2290
  %0645
  &   14.62
  &  0.6754
  &  0.1592
  &  0.1999
  %0703
  & 15.23
  &  0.7824
  &  0.2729
  &  0.1808 
  & 0.086 \\
  TensoRF   
  %0000
  & 8.94
  &  0.5894
  &  0.5878
  &  0.4021 
  %0009
  &   5.86
  &  0.1951
  &  0.8788
  &  0.5230  
  %0645
  &   7.59
  &  0.2689
  &  0.8645
  &  0.4564
  %origami
  &  7.02
  &  0.3553
  &  0.9494
  &  0.4581 
  & \yellowc0.340 \\
  Nerfacto    
  %0000
  &  8.78
  &  0.5912
  &  0.5578
  &  0.4221 
  %0009
  &   8.86
  &  0.3211
  &  0.6671
  &  0.4866  
  %0645
  &  6.95
  &  0.2997
  &  0.6585
  &  0.4546  
  %0703
  &  7.12
  &  0.3462
  &  0.8009
  &  0.4447
  & 0.132 \\
  HS-NeRF     
  %0000
  &  6.53
  &  0.3091
  &  0.2237
  &  0.4846 
  %0009
  &   8.36
  &  0.3142
  &  0.5621
  &  0.4987 
  %0645
  &  7.16
  &  0.3242
  &  0.3219
  &  0.4453 
  %0703
  &  7.40
  &  0.3558
  &  0.5351
  &  0.4310
  & 0.0642 \\
  \midrule
  3DGS   
  %0000
  & \orangec23.08 & \orangec0.8842 & \orangec0.0194 & \orangec0.0756
  %0009 
  & \orangec20.96 & \orangec0.8172 & \orangec0.0625 & \orangec0.1273
  %0645
  & \orangec17.47 & \orangec0.7435 & \orangec0.0984 & \orangec0.1539
  %0703
  & \orangec20.96 & \orangec0.8448 & \orangec0.0765 & \orangec0.09921
  & \redc81.1 \\
  HyperGS    
  %0000
  &  \redc23.11
  &  \redc0.9096
  &  \redc0.0193
  &  \redc0.0713
  %0009
  &  \redc27.20
  &  \redc0.9372
  &  \redc0.0192
  &  \redc0.0454
  %0645
  &  \redc24.03
  &  \redc0.8788
  &  \redc0.0416
  &  \redc0.0653
  %0703
  &  \redc26.13
  &  \redc0.8461
  &  \redc0.0750
  &  \redc0.0513 
  & \orangec3.11 \\
  \bottomrule
\end{tabular}

%% file: figures/tables/ablation_pruning_f.tex
\begin{tabular}{l ccccc} % 6 columns: 1 for ablation step, 5 for metrics including num primitives
  \toprule
  \multirow{2}{*}{Ablation Step} & \multicolumn{5}{c}{Average Results for Bayspec} \\ % Span over the metrics
   % Rule under the metrics
            & \PSNR & \SSIM & \MRAE & \RMSE & N.Prim(k) $\downarrow$ \\
  \cmidrule(lr){1-1} \cmidrule(lr){2-6} % Rule under the Ablation Step column
  None         & 26.68 & 0.753  & 0.0340  & 0.0442  & 1301 \\
  MSE         & 24.11 & 0.712  & 0.0340  & 0.0493  & \redc121 \\
  Huber       & \orangec27.04 & \orangec0.7742  & \orangec0.0257  & 0.0461 & \orangec218 \\
  MAE         & \yellowc27.00 & \yellowc0.7753  & \yellowc0.0269  & \yellowc0.0451  & 532 \\
  SAM         & 26.89 & 0.7651  & 0.0269  & \orangec0.0447  & 270 \\
  L1          & \redc27.11 & \redc0.7804  & \redc0.0254  & \redc0.0440 & \yellowc226 \\
  \bottomrule
\end{tabular}

%% file: figures/tables/ablation_AE_dataset_V_scene.tex
\begin{tabular}{l cccc} % 5 columns: 1 for method and 4 for metrics
  \toprule
  \multirow{2}{*}{AE type} & \multicolumn{4}{c}{Average results for Bayspec dataset} \\ % Span over the metrics
  
            & \PSNR & \SSIM & \MRAE & \RMSE \\ \cmidrule(lr){1-1} \cmidrule(lr){2-5}
  % \midrule
  General  & 26.61 & 0.7722  & 0.02791  & 0.04589 \\
  Per Scene  & 27.11 & 0.7804  & 0.0254  & 0.0440 \\
  \bottomrule
\end{tabular}

%% file: figures/tables/ablation_latent_space.tex
\begin{tabular}{cccccccc}
\hline
\multicolumn{2}{c}{\multirow{2}{*}{\begin{tabular}[c]{@{}c@{}}C.\\ Depth\end{tabular}}}  & \multicolumn{4}{c}{Average}                                                                & \multirow{2}{*}{\begin{tabular}[c]{@{}c@{}}Size\\ (GB)\end{tabular}} \\
\multicolumn{2}{c}{}                                                                     & \PSNR & \SSIM & \MRAE & \RMSE &                                                                                             \\
\cmidrule(lr){1-2} \cmidrule(lr){3-6} \cmidrule(lr){7-7} \cmidrule(lr){8-8} % Separate cmidrules for each section
\multirow{3}{*}{\rotatebox{90}{B.Spec}}  & 36  & 27.11 & 0.7804 & 0.0254 & 0.0440  & 1.27 \\
                                         & 27  & 26.11 & 0.7347 & 0.0294 & 0.0481  & 1.15 \\\hdashline
                                         % & 18  & 18.789 & 0.868 & 0.089 & 0.05 & 60 \\ \hdashline
\multirow{3}{*}{\rotatebox{90}{SOP}}     & 32 & 30.51 & 0.9756 & 0.0415 & 0.0354  & 1.24 \\
                                         & 24  & 29.21 & 0.9701 & 0.0321 & 0.0469 & 1.13 \\
\bottomrule
                                         % & 16  & 18.789 & 0.868 & 0.089 & 0.05 & 1.17 \\ \hdashline
% \multirow{3}{*}{\rotatebox{90}{S.Net}}   & 57  & 18.789 & 0.868 & 0.089 & 0.05 & 40 \\
%                                          & 42  & 18.789 & 0.868 & 0.089 & 0.05  & 45 \\
%                                          & 27  & 18.789 & 0.868 & 0.089 & 0.05 & 50 \\ \hline
\end{tabular}

%% file: figures/tables/ablation_pruning_freq.tex
\begin{tabular}{l ccccc} % 6 columns: 1 for ablation step, 5 for metrics including num primitives
  \toprule
  \multirow{2}{*}{Ablation Step} & \multicolumn{5}{c}{Average results on the Bayspec dataset} \\ % Span over the metrics
   % Rule under the metrics
            & \PSNR & \SSIM & \MRAE & \RMSE & N.Prim(k)$\downarrow$ \\
  \cmidrule(lr){1-1} \cmidrule(lr){2-6} % Rule under the Ablation Step column
  In Densif., 1    & 27.11 & 0.7804  & 0.0254  & 0.0440 & 226 \\
  In Densif, 2    & 26.97 & 0.7793  & 0.0268  & 0.0451  & 186 \\ \hdashline
  % In Densif, 3    & 19.234 & 0.860  & 0.090  & 0.048 & 1400 \\
  Post Densif., 1     & 26.11 & 0.7462  & 0.0274  & 0.0498 & 159 \\ \hdashline
  % Post Densif., 2     & 19.987 & 0.885  & 0.075  & 0.044 & 1250 \\
  % Post Densif., 3     & 19.987 & 0.885  & 0.075  & 0.044 & 1250 \\
  Hybrid, 1 & 26.09 & 0.7451  & 0.0281  & 0.0497 & 158 \\
  % Hybrid, 2 & 19.567 & 0.870  & 0.085  & 0.046 & 1300 \\
  \bottomrule
\end{tabular}